\documentclass[letterpaper]{article} 
\usepackage{aaai2026}  
\usepackage{times}  
\usepackage{helvet}  
\usepackage{courier}  
\usepackage[hyphens]{url}  
\usepackage{graphicx} 
\usepackage{natbib}  
\usepackage{caption} 
\usepackage{algorithm}
\usepackage{algorithmic}

\usepackage{algorithm}
\usepackage{algorithmic}

\usepackage{graphicx}
\usepackage{tikz}
\usepackage{color}
\usepackage{multirow}
\usepackage{booktabs}
\usepackage{amsthm}
\usepackage{amssymb}
\usepackage{amsmath}
\usepackage{tabularx, multirow, multicol, graphicx, makecell, enumitem, color, soul, colortbl, }

\usepackage{appendix}

\usepackage{newfloat}
\usepackage{listings}
\DeclareCaptionStyle{ruled}{labelfont=normalfont,labelsep=colon,strut=off} 
\floatstyle{ruled}
\newfloat{listing}{tb}{lst}{}
\floatname{listing}{Listing}
\title{HD$^2$-SSC: High-Dimension High-Density Semantic Scene Completion \\ for Autonomous Driving}

\author{
    Zhiwen Yang\textsuperscript{\rm 1},
    Yuxin Peng\textsuperscript{\rm 1}\thanks{Corresponding author.}
}
\affiliations{
    \textsuperscript{\rm 1}Wangxuan Institute of Computer Technology, Peking University\\
    \{yangzhiwen, pengyuxin\}@pku.edu.cn
}

\begin{document}

\maketitle

\begin{abstract}
Camera-based 3D semantic scene completion (SSC) plays a crucial role in autonomous driving, enabling voxelized 3D scene understanding for effective scene perception and decision-making. 
Existing SSC methods have shown efficacy in improving 3D scene representations, but suffer from the inherent input-output \textit{dimension gap} and annotation-reality \textit{density gap}, where the 2D planar view from input images with sparse annotated labels leads to inferior prediction of real-world dense occupancy with a 3D stereoscopic view.
In light of this, we propose the corresponding \textbf{H}igh-\textbf{D}imension \textbf{H}igh-\textbf{D}ensity Semantic Scene Completion (\textbf{HD$^2$-SSC}) framework with expanded pixel semantics and refined voxel occupancies.
To bridge the dimension gap, a High-dimension Semantic Decoupling module is designed to expand 2D image features along a pseudo third dimension, decoupling coarse pixel semantics from occlusions, and then identify focal regions with fine semantics to enrich image features.
To mitigate the density gap, a High-density Occupancy Refinement module is devised with a ``detect-and-refine" architecture to leverage contextual geometric and semantic structures for enhanced semantic density with the completion of missing voxels and correction of erroneous ones.
Extensive experiments and analyses on the SemanticKITTI and SSCBench-KITTI-360 datasets validate the effectiveness of our HD$^2$-SSC framework.
\end{abstract}

\begin{links}
    \link{Code}{https://github.com/PKU-ICST-MIPL/HD2-AAAI2026}
\end{links}

\begin{figure}[t]
  \includegraphics[width=1.0\columnwidth]{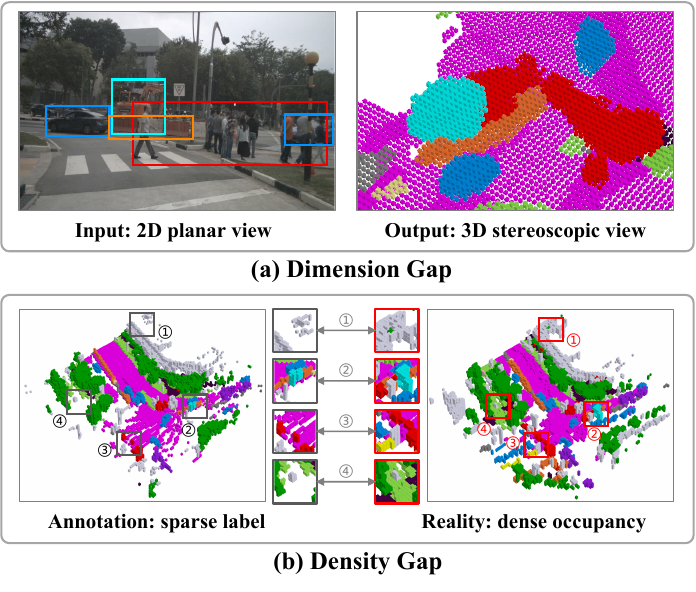}
  \caption{Illustrations of (a) Dimension Gap, highlighting the disparity between input coarse pixel semantics with occlusion and output distinct fine voxel semantics, (b) Density Gap, depicting the difference between annotated sparse labels and ground-truth dense occupancies.}
  \label{fig:motivation}
\end{figure}

\section{Introduction}
Accurate 3D geometry perception of the surrounding environment is essential for autonomous driving systems, enabling downstream tasks such as planning, navigation, and interaction with dynamic environments. 3D Semantic Scene Completion (SSC)~\cite{roldao20223d} aims to infer both the occupancy and semantics of the 3D space, providing a voxelized understanding of the scene. Considering the complexity of outdoor environments, LiDAR-based methods~\cite{roldao2020lmscnet, cheng2021s3cnet, xia2023scpnet, li2023lode, lee2024semcity, wang2025l2cocc} have been widely explored due to their ability to directly capture 3D structural information from input point clouds. However, they face the inherent challenges of high sensor cost and limited scalability for LiDAR data, restricting their applications to large-scale deployment in real-world scenarios.

Recently, camera-based semantic scene completion has garnered increasing attention as a cost-effective and scalable alternative utilizing 2D images as input. The pioneering work MonoScene~\cite{cao2022monoscene} lifts 2D image features into dense 3D volumes with depth projection. Building on this, bird's-eye-view (BEV)~\cite{huang2021bevdet, li2023bevstereo} and tri-perspective view (TVP)~\cite{huang2023tri} representations are designed for improved computation efficiency. With these advances, transformer-based methods~\cite{li2022bevformer, wei2023surroundocc} utilized voxel queries to aggregate image features into 3D scenes. VoxFormer~\cite{li2023voxformer} introduces a two-stage framework, aggregating 3D features of visible areas and then diffusing to the whole scene. Efforts have been made for improved performance through neighboring object structures and view spatial differences~\cite{xue2024bi}, context-dependent queries~\cite{yu2024context}, and dense-sparse-dense design for hybrid guidance~\cite{mei2024camera}.

Despite significant progress, existing methods typically focus on refining voxel-based scene representations, but treat pixel features and voxel semantics indiscriminately during view transformation and occupancy prediction, facing two key challenges. (1) \textbf{Dimension Gap}: As illustrated in Figure \ref{fig:motivation} (a), input images are captured from a 2D planar view, leading to coarse pixel features confused by multiple object semantics with occlusions. In contrast, SSC requires fine-grained voxel semantics for distinct objects in a 3D stereoscopic view, necessitating the expansion and decoupling of coarse pixel features from occlusion. (2) \textbf{Density Gap}: As shown in Figure \ref{fig:motivation} (b), manual annotations derived from LiDAR sensors yield inherently sparse labels with interspace due to the finite resolution of collected point clouds. On the other hand, real-world scenes exhibit dense, consistent occupancy with rich contextual details, demanding comprehensive detection and meticulous refinement of voxel occupancy to enrich semantic density.

To address these challenges, we introduce the \textit{High-Dimension High-Density Semantic Scene Completion} (\textbf{HD$^2$-SSC}) framework. Specifically, the \textit{High-dimension Semantic Decoupling} (\textbf{HSD}) module first expands and decouples coarse pixel semantics to alleviate the dimension gap. 
The process begins with expanding 2D image features into voxelized features through the Pseudo Voxelization block, with an orthogonal loss promoting distinct expanded semantics. The Semantic Aggregation block then extracts global semantics with pixel queries and integrates high-dimension voxelized features using semantic clustering with decoupling loss.
Then, to mitigate the density gap, the \textit{High-density Occupancy Refinement} (\textbf{HOR}) module operates in a ``detect-and-refine" manner to enhance semantic density with contextual geometric and semantic structures. In the detection phase, binary predictions are generated for a comprehensive collection of occupied voxels, focusing on the identification of geometric critical voxels. In the subsequent refinement phase, we conduct class-wise prediction to identify semantic critical voxels, with respect to the prediction confidence score. Then the overall distributions of geometric and semantic critical voxels are aligned for consistent contextual structures, enhancing semantic density with completion of missing voxels and correction of erroneous ones.
Extensive experiments and analyses on SemanticKITTI and SSCBench-KITTI-360 demonstrate the superiority of our HD$^2$-SSC over SOTA methods.

The main contributions are summarized as follows:
\vspace{-\topsep}
\begin{itemize}
    \item[$\bullet$]
    We propose HD$^2$-SSC to address the challenges of dimension and density gap in camera-based semantic scene completion, improving SSC performance with completion of missing voxels and correction of erroneous ones. 
    \item[$\bullet$]
    HSD bridges the dimension gap through expanding and decoupling coarse pixel semantics with orthogonal loss, then aggregating high-dimension voxelized semantics upon semantic clustering with decoupling loss.
    \item[$\bullet$] 
    HOR addresses the density gap in a ``detect-and-refine" paradigm to identify geometric and semantic critical voxels, aligning the overall critical distributions for consistent contextual details and improved semantic density.
\end{itemize}\par 

\begin{figure*}[ht]
\centering
\includegraphics[width=1.0\textwidth]{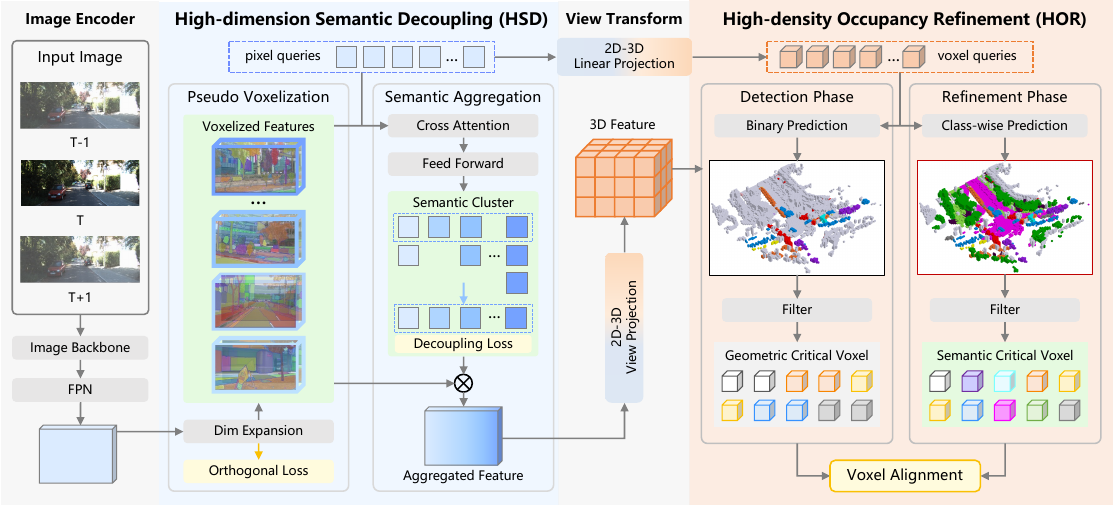}
\caption{The overall architecture of our HD$^2$-SSC. The High-dimension Semantic Decoupling (HSD) module expands and decouples coarse pixel semantics with orthogonal loss, then aggregates high-dimension voxelized semantics via semantic clustering with decoupling loss. The High-density Occupancy Refinement (HOR) module adopts a ``detect-and-refine" architecture to identify geometric and semantic critical voxels, whose overall distributions are aligned for consistent contextual details.}
\label{framework}
\end{figure*}

\section{Related Work}

\subsection{LiDAR-based SSC Methods}
Due to the complexity of outdoor scenes and rich 3D structural information embedded in point clouds, LiDAR-based methods~\cite{liong2020amvnet, cheng20212, zhang2023attention, ye2023lidarmultinet} have long been the predominant solutions for SSC. UDNet~\cite{zou2021up} pioneered the use of a 3D U-Net architecture to generate scene predictions directly from LiDAR data. SGCNet~\cite{zhang2018efficient} enhanced efficiency through spatial group convolutions for sparse 3D processing. Further advancements have focused on refining 3D scene representations, such as multi-view fusion for completing sparse scenes~\cite{cheng2021s3cnet} and local implicit functions for continuous scene modeling~\cite{rist2021semantic}. SCPNet~\cite{xia2023scpnet} transferred multi-frame information to single-frame models for lightweight architectures. SSA-SC~\cite{yangsemantic} and SSC-RS~\cite{mei2023ssc} adopted multi-branch designs to hierarchically fuse semantic and geometric features. Despite the advancements, LiDAR-based methods remain computationally expensive due to large volume of LiDAR points, limiting their scalability to real-world, large-scale applications.

\subsection{Camera-based SSC Methods}
In recent years, camera-based SSC methods~\cite{miao2023occdepth, yao2023ndc, tong2023scene, wang2024h2gformer, wang2024not, yang2025sphere, yang2025gala} have gained increasing attention due to their cost-effectiveness and ease of deployment.
One of the first fully visual SSC frameworks, MonoScene~\cite{cao2022monoscene}, pioneers a voxel-based approach by sampling image features along lines of sight. Another line of research leverages Bird's-Eye-View (BEV) representations to aggregate multi-view image features for 3D scene modeling. LSS~\cite{philion2020lift} introduces a frustum-based lifting mechanism, which is further enhanced by BEVDet~\cite{huang2021bevdet}, aligning multi-frame features for improved scene comprehension. BEVDepth~\cite{li2023bevdepth} and BEVStereo~\cite{li2023bevstereo} incorporate camera-aware depth estimation and dynamic temporal stereo information for more precise depth learning. Beyond BEV, TPVFormer~\cite{huang2023tri} proposes a Tri-Perspective View (TPV) representation, decomposing voxel grids onto three orthogonal planes for efficient scene encoding. Transformer-based methods utilize BEV queries~\cite{jiang2023polarformer} or voxel queries~\cite{ma2024cotr}. Voxformer~\cite{li2023voxformer} adopts a two-stage framework with class-agnostic query proposals. Symphonize~\cite{jiang2024symphonize} introduces a scene-from-instance paradigm to enhance feature interactions. HASSC~\cite{wang2024not} exploits the local and global hardness of each voxel, together with a self-distillation strategy for stable and consistent training process. SGN~\cite{mei2024camera} presents a one-stage SSC framework with a dense-sparse-dense strategy, exploiting hybrid guidance to improve segmentation boundaries for more precise 3D semantic scene completion.

In general, existing camera-based SSC methods mainly focus on devising delicate architectures to refine 3D features, but overlook two critical challenges: the input-output dimension gap and annotation-reality density gap. In contrast, our HD$^2$-SSC approach decouples coarse pixel semantics to bridge the dimension gap for aligned view transformation, and aligns critical voxel distributions for consistent contextual details to mitigate the density gap.

\section{Approach}
Figure~\ref{framework} illustrates our High-Dimension High-Density 3D Semantic Scene Completion (HD$^2$-SSC) framework, consisting of four main modules: (1) an Image Encoder extracting 2D features, (2) a High-dimension Semantic Decoupling (HSD) module that addresses the dimension gap by decoupling coarse pixel semantics with orthogonal loss and aggregating high-dimension voxelized semantics upon semantic clustering, (3) a View Transform module projecting 2D pixel features to 3D voxel features, and (4) a High-density Occupancy Refinement (HOR) module that mitigates the density gap through aligning geometric and semantic critical voxel distributions in a ``detect-and-refine" manner.

\subsection{Preliminary}
\subsubsection{Problem Setup.} Given input stereo images $I_{l}^{\rm rgb}, I_{r}^{\rm rgb}$, the objective of SSC is to predict the geometry and semantics of 3D scenes in front. The output is represented as a voxel grid $Y\in\mathbb{R}^{H\times W\times Z}$, where $H, W, Z$ correspond to the grid's length, width, and height, respectively. Each voxel is categorized as either empty denoted by $c_0$ or occupied by one of the semantic classes in $c\in\{c_1,\cdots,c_N\}$, where $N$ represents the number of semantic classes.
\subsubsection{Image Encoder.} Consistent with previous methods~\cite{mei2024camera}, we employ the ResNet-50~\cite{he2016deep} network with FPN~\cite{lin2017feature} to extract 2D features $F^{\rm 2D}\in\mathbb{R}^{N_t\times C\times H\times W}$ from input RGB images, where $N_t$ is the image number of temporal inputs, $C$ is the feature channel and $(H,W)$ denotes the image resolution.
\subsubsection{View Transformer.} Following~\cite{mei2024camera}, we construct 3D features by sampling 2D features via 3D-2D projection mapping with camera parameters. Let $x\in\mathbb{R}^{H\times W\times Z\times 3}$ denote the centroid of $H\times W\times Z$ voxels in world coordinates. We establish the projection mapping $\pi(x)$ from pixel $(u, v)$ using the camera intrinsic parameter $K$ and extrinsic parameter $T=[R,t]$:
\begin{equation}
\begin{aligned}
    [h_c,w_c,z_c]^{T}&=R\cdot p+t\\
    z_c\circ[u,v,1]^{T}&=K\cdot[h_c,w_c,z_c]^{T}
\end{aligned}
\end{equation}
where $\circ$ denotes element-wise product.
\subsection{High-dimension Semantic Decoupling}
The High-dimension Semantic Decoupling (HSD) module consists of two blocks for addressing the dimension gap. 
The Pseudo Voxelization block expands and decouples coarse pixel semantics, and the Semantic Aggregation block aggregates the high-dimension voxelized semantics.
\subsubsection{Pseudo Voxelization.} Given image feature \( F_{\rm cam} \) extracted from the image encoder, a Dim Expansion (DE) layer is employed to lift it along a pseudo ``semantic dimension" into pseudo voxelized features. This step is essential for addressing the dimension gap, generating multiple candidates for decoupling occluded objects within coarse pixel semantics: 
\begin{equation}
    F_{\rm pseudo} = {\rm DE}(F_{\rm cam}, D_{\rm exp})
\end{equation}
where $F_{\rm pseudo}=\{F_{\rm pseudo}^{i}\in\mathbb{R}^{N_{\rm cam}\times C_{\rm 2D} \times H _{\rm 2D}\times W_{\rm 2D}}\}_{i=1}^{D_{\rm exp}}$ represents pseudo voxelized features, and $D_{\rm exp}$ is the channel of expanded dimension, ${\rm DE}(\cdot)$ represents the dim expansion operation with 2D convolution layers.
However, the pseudo voxelized features may still exhibit similar semantics at the same 2D coordinates \( (h_{\rm 2D}, w_{\rm 2D}) \), which limits their ability to generalize across different occluded objects. Therefore, we employ an orthogonal loss to further promote the distinct voxelized semantics:
\begin{equation}
    \mathcal{L}_{\rm orth} = \lambda\left|W_{\rm DE}W_{\rm DE}^{\rm T} - I\right|
\end{equation}
where $W_{\rm DE}$ represents the weight matrix of the dim expansion projection, $I$ indicates the identity matrix, and $\lambda$ is the regularization parameter for the orthogonal loss.

\subsubsection{Semantic Aggregation.}
We first introduce pixel queries $Q_{\rm pixel}\in\mathbb{R}^{N_{\rm query}\times C_{\rm 2D}}$ to collect global semantics from the pseudo voxelized features through cross attention mechanism. 
Then, to further assign different pseudo voxelized feature slices $F_{\rm pseudo}^{i}$ with discrete semantics, we perform semantic clustering (DPC-kNN) on the collected global semantics in $Q_{\rm pixel}$:
\begin{equation}
    {\rm DPC}(Q_{\rm pixel}, D_{\rm exp}) = \mathop{\arg\min}\limits_{C}\sum\limits_{i=1}^{D_{\rm exp}}\sum\limits_{v\in C_{i}}||v-c_i||^2
\end{equation}
where \( C \) represents the \( D_{\rm exp} \) semantic clusters corresponding to the expanded dimension of the pseudo voxelized features, and \( c_i \) is the centroid of the \( i \)-th cluster \( C_i \). Additionally, a decoupling loss is adopted to enhance the distinction among different semantic clusters:
\begin{equation}
    \mathcal{L}_{\rm decouple} = \sum\limits_{i\neq j}\dfrac{C_i\cdot C_j}{|C_i|\cdot|C_j|}
\end{equation}

\begin{figure}[t]
\centering
\includegraphics[width=1.0\columnwidth]{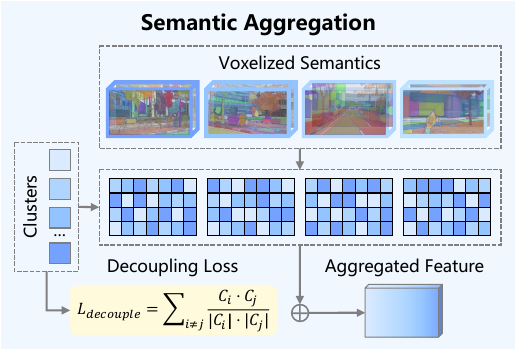} 
\caption{Illustration of aggregating the high-dimension voxelized semantics concerning the semantic clusters with decoupling loss.}
\label{fig:decouple}
\end{figure}

Then, discriminative regions of the pseudo voxelized features are located with respect to the similarity against semantic clusters, thereby aggregating high-dimension voxelized semantics, as illustrated in Figure~\ref{fig:decouple}:
\begin{equation}
    F_{\rm agg}=\sum\limits_{i=1}^{D_{\rm exp}}F_{\rm pseudo}^{i}\cdot\mathop{\max}\limits_{C_j} {\rm sim}(F_{\rm pseudo}^{i}, C_{j})
\end{equation}

\subsection{High-density Occupancy Refinement}
The High-density Occupancy Refinement (HOR) module is designed in a ``detect-and-refine" architecture to identify geometric and semantic critical voxels, respectively. Then the critical voxels are aligned to ensure contextual geometric and semantic consistency.
\begin{table*}
  \centering
  \resizebox{1.0\linewidth}{!}{
  \begin{tabular}{l|cc|ccccccccccccccccccc}
    \toprule
    \multirow{2}{*}{Method} & SC & SSC
    & \rotatebox{90}{\multirow{2}{*}{\textcolor[RGB]{91,155,213}{$\blacksquare$} \textbf{car}{\footnotesize (3.92\%)}}} 
    & \rotatebox{90}{\multirow{2}{*}{\textcolor[RGB]{100,230,245}{$\blacksquare$} \textbf{bicycle}{\footnotesize (0.03\%)}}} 
    & \rotatebox{90}{\multirow{2}{*}{\textcolor[RGB]{30,60,150}{$\blacksquare$} \textbf{motorcycle}{\footnotesize (0.03\%)}}} 
    & \rotatebox{90}{\multirow{2}{*}{\textcolor[RGB]{80,30,180}{$\blacksquare$} \textbf{truck}{\footnotesize (0.16\%)}}}
    & \rotatebox{90}{\multirow{2}{*}{\textcolor[RGB]{0,0,255}{$\blacksquare$} \textbf{other-veh.}{\footnotesize (0.20\%)}}}
    & \rotatebox{90}{\multirow{2}{*}{\textcolor[RGB]{255,30,30}{$\blacksquare$} \textbf{person}{\footnotesize (0.07\%)}}}
    & \rotatebox{90}{\multirow{2}{*}{\textcolor[RGB]{255,37,199}{$\blacksquare$} \textbf{bicyclist}{\footnotesize (0.07\%)}}}
    & \rotatebox{90}{\multirow{2}{*}{\textcolor[RGB]{150,30,90}{$\blacksquare$} \textbf{motorcyclist}{\footnotesize (0.05\%)}}}
    & \rotatebox{90}{\multirow{2}{*}{\textcolor[RGB]{255,0,255}{$\blacksquare$} \textbf{road}{\footnotesize (15.30\%)}}}
    & \rotatebox{90}{\multirow{2}{*}{\textcolor[RGB]{255,150,255}{$\blacksquare$} \textbf{parking}{\footnotesize (1.12\%)}}}
    & \rotatebox{90}{\multirow{2}{*}{\textcolor[RGB]{75,0,75}{$\blacksquare$} \textbf{sidewalk}{\footnotesize (11.13\%)}}}
    & \rotatebox{90}{\multirow{2}{*}{\textcolor[RGB]{175,0,75}{$\blacksquare$} \textbf{other-grnd.}{\footnotesize (0.56\%)}}}
    & \rotatebox{90}{\multirow{2}{*}{\textcolor[RGB]{255,200,0}{$\blacksquare$} \textbf{building}{\footnotesize (14.10\%)}}}
    & \rotatebox{90}{\multirow{2}{*}{\textcolor[RGB]{255,120,50}{$\blacksquare$} \textbf{fence}{\footnotesize (3.90\%)}}}
    & \rotatebox{90}{\multirow{2}{*}{\textcolor[RGB]{0,175,0}{$\blacksquare$} \textbf{vegetation}{\footnotesize (39.3\%)}}}
    & \rotatebox{90}{\multirow{2}{*}{\textcolor[RGB]{135,60,0}{$\blacksquare$} \textbf{trunk}{\footnotesize (0.51\%)}}}
    & \rotatebox{90}{\multirow{2}{*}{\textcolor[RGB]{150,240,80}{$\blacksquare$} \textbf{terrain}{\footnotesize (9.17\%)}}}
    & \rotatebox{90}{\multirow{2}{*}{\textcolor[RGB]{255,240,150}{$\blacksquare$} \textbf{pole}{\footnotesize (0.29\%)}}}
    & \rotatebox{90}{\multirow{2}{*}{\textcolor[RGB]{255,0,0}{$\blacksquare$} \textbf{traf.-sign}{\footnotesize (0.08\%)}}}\\
     & IoU & mIoU & & & & & & & & & & & & & & & & & & &\\
     \midrule
     \multicolumn{22}{l}{\textit{Mono-Input Methods}} \\
    \midrule
    MonoScene$^{\dagger}$~\cite{cao2022monoscene} & 36.86 & 11.08 & 23.26 & 0.61 & 0.45 & 6.98 & 1.48 & 1.86 & 1.20 & 0.00 & 56.52 & 14.27 & 26.72 & 0.46 & 14.09 & 5.84 & 17.89 & 2.81 & 29.64 & 4.14 & 2.25 \\
    TPVFormer$^{\dagger}$~\cite{huang2023tri} & 35.61 & 11.36 & 23.81 & 0.36 & 0.05 & 8.08 & 4.35 & 0.51 & 0.89 & 0.00 & 56.50 & 20.60 & 25.87 & 0.85 & 13.88 & 5.94 & 16.92 & 2.26 & 30.38 & 3.14 & 1.52 \\
    OccFormer$^{\dagger}$~\cite{zhang2023occformer} & 36.50 & 13.46 & 25.09 & 0.81 & 1.19 & 25.53 & 8.52 & 2.78 & 2.82 & 0.00 & 58.85 & 19.61 & 26.88 & 0.31 & 14.40 & 5.61 &  19.63 & 3.93 & 32.62 & 4.26 & 2.86\\
    IAMSSC$^{*}$~\cite{xiao2024instance} & 44.29 & 12.45 & 26.26 & 0.60 & 0.15 & 8.74 & 5.06 & 1.32 & 3.46 & 0.01 & 54.55 & 16.02 & 25.85 & 0.70 & 17.38 & 6.86 & 24.63 & 4.95 & 30.13 & 6.35 & 3.56\\
    \midrule
    \multicolumn{22}{l}{\textit{Temporal-Input Methods}} \\
    \midrule
    VoxFormer$^{*}$~\cite{li2023voxformer} & 44.15 & 13.35 & 26.54 & 1.28 & 0.56 & 7.26 & 7.81 & 1.93 & 1.97 & 0.00 & 53.57 & 19.69 & 26.52 & 0.42 & 19.54 & 7.31 & 26.10 & 6.10 & 33.06 & 9.15 & 4.94\\
    DepthSSC$^{*}$~\cite{yao2023depthssc} &45.84 & 13.28 & 25.94 & 0.35 & 1.16 & 6.02 & 7.50 & \underline{2.58} & \textbf{6.32} & 0.00 & 55.38 & 18.76 & 27.04 & \underline{0.92} & 19.23 & 8.46 & 26.37 & 4.52 & 30.19 & 7.42 & 4.09\\
    Symphonies$^{*}$~\cite{jiang2024symphonize} & 41.92 & 14.89 & 28.68 & \underline{2.54} & \textbf{2.82} & \underline{20.44} & \textbf{13.89} & \textbf{3.52} & 2.24 & 0.00 & 56.37 & 15.28 & 27.58 & 0.95 & 21.64 & 8.40 & 25.72 & 6.60 & 30.87 & 9.57 & 5.76\\
    HASSC$^{*}$~\cite{wang2024not} & 44.58 & 14.74 & 27.33 & 1.07 & 1.14 & 17.06 & 8.83 & 2.25 & \underline{4.09} & 0.00 & 57.23 & 19.89 & 29.08 & \textbf{1.26} & 20.19 & 7.95 & 27.01 & 7.71 & 33.95 & 9.20 & 4.81\\
    H2GFormer$^{*}$~\cite{wang2024h2gformer} & 44.69 & 14.29 & 28.21 & 0.95 & 0.91 & 6.80 & 9.32 & 1.15 & 0.10 & 0.00 & 57.00 & 21.74 & 29.37 & 0.34 & 20.51 & 7.98 & 27.44 & 7.80 & 36.26 & 9.88 & 5.81\\
    CGFormer$^{\dagger}$~\cite{yu2024context} & 45.99 & \underline{16.87} & \underline{34.32} & \textbf{4.61} & \underline{2.71} & \underline{19.44} & 7.67 & 2.38 & 4.08 & 0.00 & \textbf{65.51} & \underline{20.82} & \underline{32.31} & 0.16 & 23.52 & \underline{9.20} & 26.93 & 8.83 & \textbf{39.54} & 10.67 & \underline{7.84}\\
    SGN$^{*}$~\cite{mei2024camera} & \underline{46.21} & 15.32 & \underline{33.31} & 0.61 & 0.46 & 6.03 & 9.84 & 0.47 & 0.10 & 0.00 & 59.10 & 19.05 & 29.41 & 0.33 & \underline{25.17} & \textbf{9.96} & \underline{28.93} & \underline{9.58} & 38.12 & \underline{13.25} & 7.32\\
    \midrule
    \rowcolor{gray!10} 
    \textbf{HD$^2$-SSC}$^{*}$~\textbf{(Ours)} & \textbf{47.59} & \textbf{17.44} & 32.68 & 1.72 & 0.54 & \textbf{24.76} & \underline{10.73} & 1.61 & 1.73 & 0.00 & \underline{60.80} & \textbf{23.72} & \textbf{33.12} & 0.11 & \textbf{28.55} & 8.83 & \textbf{30.90} & \textbf{11.08} & \underline{38.78} & \textbf{13.41} & \textbf{8.34} \\
    \bottomrule
  \end{tabular}}
  \caption{Camera-based 3D semantic scene completion results on the SemanticKITTI~\cite{behley2019semantickitti} validation set. $^{\dagger}$ denotes the methods employing EfficientNet-B7~\cite{tan2019efficientnet} as image backbone, and $^{*}$ represents the methods utilizing ResNet-50~\cite{he2016deep} as image backbone. Best results are highlights in bold, and second-best scores are \underline{underlined}.}
  \label{kitti}
\end{table*}
\\
\subsubsection{Detection Phase.} 
During the detection phase, the voxel queries $Q_{\rm voxel}\in\mathbb{R}^{N_{\rm query}\times C_{\rm 3D}}$ and voxel feature $F_{\rm voxel}\in\mathbb{R}^{C_{\rm 3D}\times H_{\rm 3D}\times W_{\rm 3D}\times Z_{\rm 3D}}$ are passed through a binary classification head for a comprehensive detection of occupied voxels, together with a heuristic separation between foreground and background voxels:
\begin{equation}
    M_{\rm o-f}, M_{\rm f-b} = H_{\rm bc}(F_{\rm voxel}, Q_{\rm voxel})
\end{equation}
where $H_{\rm bc}$ denotes the binary classification head, $M_{\rm o-f}, M_{\rm f-b}$ represent the score maps distinguishing occupied voxels from free ones, foreground voxels from background ones, respectively. The generated score maps are added to represent the geometric density score of each voxel, indicating its general structural significance. Then we select geometric critical voxels with the highest geometric density scores:
\begin{equation}
    V_{\rm geo} = {\rm top}_{k}(M_{\rm o-f} + M_{\rm f-b})
\end{equation}
where $k$ is the number of selected geometric critical voxels.

In the detection phase, we focus on the general classification between occupied and free, foreground and background voxels, and select geometric critical voxels with discriminative structural contextual information.

\noindent
\subsubsection{Refinement Phase.} During the refinement phase, we first conduct class-wise prediction with the voxel queries and voxel features to generate the initial semantic scene completion results:
\begin{equation}
    Y_{\rm init} = H_{\rm cc}(F_{\rm voxel}, Q_{\rm voxel})
\end{equation}
where $H_{\rm cc}$ denotes the class-wise classification head, $Y_{\rm init}$ represents the initial semantic scene completion results. Furthermore, we select semantic critical voxels with $Y_{\rm init}$ indicating the classification confidence and contextual semantic significance:
\begin{equation}
    V_{\rm sem} = {\rm top}_{k}\left(\max(Y_{\rm init})\right)
\end{equation}
where $k$ is the number of selected semantic critical voxels, aligned with geometric critical voxels.
\\
\subsubsection{Voxel Alignment.} After selecting geometric and semantic critical voxels with discriminative geometric and semantic contextual information, respectively, we align the overall critical voxel distributions to promote consistent contextual semantic and geometry for further refinement:
\begin{equation}
    \mathcal{L}_{\rm critical} = \mathcal{L}_{\rm KL}(V_{\rm geo} || V_{\rm sem}) + \mathcal{L}_{\rm KL}(V_{\rm sem} || V_{\rm geo})
\end{equation}
where $\mathcal{L}_{\rm KL}(\cdot || \cdot)$ is the Kullback-Leibler divergence loss measuring the difference between two distributions. With the aligned critical voxels serving as soft constraint on the complementary effect of contextual geometry and semantics, we utilize a refined multi-layer perceptron (${\rm MLP_{refine}}$) to enhance the initial SSC predictions:
\begin{equation}
    Y_{\rm refine} = Y_{\rm init} + {\rm MLP_{refine}}([V_{\rm geo}, V_{\rm sem}])
\end{equation}
where $Y_{\rm refine}$ represents the final refined SSC prediction, $[\cdot, \cdot]$ is the concatenation operation.

\begin{table*}
  \centering
  \resizebox{1.0\linewidth}{!}{
  \begin{tabular}{l|cc|cccccccccccccccccc}
    \toprule
    \multirow{2}{*}{Method} & SC & SSC
    & \rotatebox{90}{\multirow{2}{*}{\textcolor[RGB]{91,155,213}{$\blacksquare$} \textbf{car}{\footnotesize (2.85\%)}}} 
    & \rotatebox{90}{\multirow{2}{*}{\textcolor[RGB]{100,230,245}{$\blacksquare$} \textbf{bicycle}{\footnotesize (0.01\%)}}} 
    & \rotatebox{90}{\multirow{2}{*}{\textcolor[RGB]{30,60,150}{$\blacksquare$} \textbf{motorcycle}{\footnotesize (0.01\%)}}} 
    & \rotatebox{90}{\multirow{2}{*}{\textcolor[RGB]{80,30,180}{$\blacksquare$} \textbf{truck}{\footnotesize (0.16\%)}}}
    & \rotatebox{90}{\multirow{2}{*}{\textcolor[RGB]{0,0,255}{$\blacksquare$} \textbf{other-veh.}{\footnotesize (5.75\%)}}}
    & \rotatebox{90}{\multirow{2}{*}{\textcolor[RGB]{255,30,30}{$\blacksquare$} \textbf{person}{\footnotesize (0.02\%)}}}
    & \rotatebox{90}{\multirow{2}{*}{\textcolor[RGB]{255,0,255}{$\blacksquare$} \textbf{road}{\footnotesize (14.98\%)}}}
    & \rotatebox{90}{\multirow{2}{*}{\textcolor[RGB]{255,150,255}{$\blacksquare$} \textbf{parking}{\footnotesize (2.31\%)}}}
    & \rotatebox{90}{\multirow{2}{*}{\textcolor[RGB]{75,0,75}{$\blacksquare$} \textbf{sidewalk}{\footnotesize (6.43\%)}}}
    & \rotatebox{90}{\multirow{2}{*}{\textcolor[RGB]{175,0,75}{$\blacksquare$} \textbf{other-grnd.}{\footnotesize (2.05\%)}}}
    & \rotatebox{90}{\multirow{2}{*}{\textcolor[RGB]{255,200,0}{$\blacksquare$} \textbf{building}{\footnotesize (15.67\%)}}}
    & \rotatebox{90}{\multirow{2}{*}{\textcolor[RGB]{255,120,50}{$\blacksquare$} \textbf{fence}{\footnotesize (0.96\%)}}}
    & \rotatebox{90}{\multirow{2}{*}{\textcolor[RGB]{0,175,0}{$\blacksquare$} \textbf{vegetation}{\footnotesize (41.99\%)}}}
    & \rotatebox{90}{\multirow{2}{*}{\textcolor[RGB]{150,240,80}{$\blacksquare$} \textbf{terrain}{\footnotesize (7.10\%)}}}
    & \rotatebox{90}{\multirow{2}{*}{\textcolor[RGB]{255,240,150}{$\blacksquare$} \textbf{pole}{\footnotesize (0.22\%)}}}
    & \rotatebox{90}{\multirow{2}{*}{\textcolor[RGB]{255,0,0}{$\blacksquare$} \textbf{traf.-sign}{\footnotesize (0.06\%)}}}
    & \rotatebox{90}{\multirow{2}{*}{\textcolor[RGB]{250,150,0}{$\blacksquare$} \textbf{other-struct.}{\footnotesize (4.33\%)}}}
    & \rotatebox{90}{\multirow{2}{*}{\textcolor[RGB]{50,255,255}{$\blacksquare$} \textbf{other-obj.}{\footnotesize (0.28\%)}}}\\
     & IoU & mIoU & & & & & & & & & & & & & & & & & &\\
     \midrule
     \multicolumn{21}{l}{\textit{Mono-Input Methods}} \\
    \midrule
    MonoScene$^{\dagger}$~\cite{cao2022monoscene} & 37.87 & 12.31 & 19.34 & 0.43 & 0.58 & 8.02 & 2.03 & 0.86 & 48.35 & 11.38 & 28.13 & 3.32 & 32.89 & 3.53 & 26.15 & 16.75 & 6.92 & 5.67 & 4.20 & 3.09\\
    GaussianFormer2$^{*}$~\cite{huang2024probabilistic} & 38.31 & 13.90 & 21.08 & 2.55 & 4.21 & 12.41 & 5.73 & 1.59 & 54.12 & 11.04 & 32.31 & 3.34 & 32.01 & 4.98 & 28.94 & 17.33 & 3.57 & 5.48 & 5.88 & 3.54\\
    TPVFormer$^{\dagger}$~\cite{huang2023tri} & 40.22 & 13.64 & 21.56 & 1.09 & 1.37 & 8.06 & 2.57 & 2.38 & 52.99 & 11.99 & 31.07 & 3.78 & 34.83 & 4.80 & 30.08 & 17.52 & 7.46 & 5.86 & 5.48 & 2.70\\
    OccFormer$^{\dagger}$~\cite{zhang2023occformer} & 40.27 & 13.81 & 22.58 & 0.66 & 0.26 & 9.89 & 3.82 & 2.77 & 54.30 & 13.44 & 31.53 & 3.55 & 36.42 & 4.80 & 31.00 & 19.51 & 7.77 & 8.51 & 6.95 & 4.60\\
    IAMSSC$^{*}$~\cite{xiao2024instance} & 41.80 & 12.97 & 18.53 & 2.45 & 1.76 & 5.12 & 3.92 & 3.09 & 47.55 & 10.56 & 28.35 & 4.12 & 31.53 & 6.28 & 29.17 & 15.24 & 8.29 & 7.01 & 6.35 & 4.19\\
    \midrule
    \multicolumn{21}{l}{\textit{Temporal-Input Methods}} \\
    \midrule
    VoxFormer$^{*}$~\cite{li2023voxformer} & 38.76 & 11.91 & 17.84 & 1.16 & 0.89 & 4.56 & 2.06 & 1.63 & 47.01 & 9.67 & 27.21 & 2.89 & 31.18 & 4.97 & 28.99 & 14.69 & 6.51 & 6.92 & 3.79 & 2.43\\
    DepthSSC$^{*}$~\cite{yao2023depthssc} & 40.85 & 14.28 & 21.90 & 2.36 & \underline{4.30} & 11.51 & 4.56 & 2.92 & 50.88 & 12.89 & 30.27 & 2.49 & 37.33 & 5.22 & 29.61 & 21.59 & 5.97 & 7.77 & 5.24 & 3.51\\
    Symphonies$^{*}$~\cite{jiang2024symphonize} & 44.12 & 18.58 & \underline{30.02} & 1.85 & \textbf{5.90} & \underline{25.07} & \underline{12.06} & \textbf{8.20} & 54.94 & 13.83 & 32.76 & \underline{6.93} & 35.11 & \underline{8.58} & 38.33 & 11.52 & 14.01 & 9.57 & \underline{14.44} & \underline{11.28}\\
    CGFormer$^{\dagger}$~\cite{yu2024context} & \underline{48.07} & \underline{20.05} & 29.85 & \underline{3.42} & 3.96 & 17.59 & 6.79 & \underline{6.63} & \textbf{63.85} & \underline{17.15} & \textbf{40.72} & 5.53 & \textbf{42.73} & 8.22 & \underline{38.80} & \textbf{24.94} & 16.24 & \textbf{17.45} & 10.18 & 6.77\\
    SGN$^{*}$~\cite{mei2024camera} & 47.06 & 18.25 & 29.03 & \textbf{3.43} & 2.90 & 10.89 & 5.20 & 2.99 & 58.14 & 15.04 & 36.40 & 4.43 & \underline{42.02} & 7.72 & 38.17 & \underline{23.22} & \underline{16.73} & \underline{16.38} & 9.93 & 5.86\\
    \midrule
    \rowcolor{gray!10} 
    \textbf{HD$^2$-SSC}$^{*}$~\textbf{(Ours)} & \textbf{48.58} & \textbf{20.62} & \textbf{33.00} & 0.00 & 2.98 & \textbf{26.90} & \textbf{12.38} & 4.75 & \underline{60.17} & \textbf{17.58} & \underline{37.55} & \textbf{8.7} & 40.72 & \textbf{9.51} & \textbf{41.69} & 14.69 & \textbf{18.7} & 12.35 & \textbf{16.65} & \textbf{12.75}\\
    \bottomrule
  \end{tabular}}
  \caption{Camera-based 3D semantic scene completion results on the SSCBench-KITTI-360~\cite{behley2019semantickitti} test set. $^{\dagger}$ denotes the methods employing EfficientNet-B7~\cite{tan2019efficientnet} as image backbone, and $^{*}$ represents the methods utilizing ResNet-50~\cite{he2016deep} as image backbone. Best results are highlights in bold, and second-best scores are \underline{underlined}.}
  \label{kitti360}
\end{table*}

\section{Experiments}
In accordance with existing camera-based SSC methods~\cite{mei2024camera, li2023voxformer}, extensive experiments and analyses are conducted to validate our HD$^2$-SSC framework on the popular SemanticKITTI~\cite{behley2019semantickitti} and SSCBench-KITTI-360~\cite{li2023sscbench} datasets.

\subsection{Experimental Setup}
\subsubsection{Dataset.} 
(1) The \textbf{SemanticKITTI}~\cite{behley2019semantickitti} dataset, based on the KITTI Odometry Benchmark~\cite{geiger2012we}, contains 22 autonomous driving sequences with 20 semantic classes and annotations. 
We follow the official protocol to split the total sequences into (00-07, 09-10) / (08) / (11-21) for the training / validation / test sets respectively. 
(2) The \textbf{SSCBench-KITTI-360}~\cite{li2023sscbench} dataset consists of 9 densely annotated sequences of urban driving scenes. It is separated into a training set from sequences (00, 02-05, 07, 10), a validation set from sequence (06), and a test set from sequence (09). For both datasets, the semantic labels are within the range of $[0\sim 51.2m, -25.6\sim25.6m, -2\sim4.4m]$, and the target voxel grids are represented as $256 \times 256 \times 32$ voxel grids with the resolution of 0.2$m$.
\noindent
\subsubsection{Evaluation Metrics.}
Following~\cite{mei2024camera}, we adopt the Intersection over Union (IoU) of occupied voxels as the evaluation metric for the class-agnostic scene completion (SC) task, and the mean Intersection over Union (mIoU) metric for the semantic scene completion (SSC) task.
\begin{table}[t]
    \centering
    \begin{tabular}{l|c|c}
    \toprule
    \textbf{Variants} & \textbf{IoU} & \textbf{mIoU} \\
    \midrule
    \midrule
    Baseline & 44.15 & 13.35 \\
    Baseline + HSD & 46.45 & 15.58 \\
    Baseline + HOR & 46.07 & 16.12 \\
    \rowcolor{gray!10} 
    Baseline + HSD + HOR (\textbf{Ours}) & \textbf{47.59} & \textbf{17.44} \\
    \bottomrule
    \end{tabular}
    \caption{Ablation study on the SemanticKITTI validation set dataset of different components of our HD$^2$-SSC.}
    \label{ablation}
\end{table}
\\
\begin{table}[t]
    \centering
    \begin{tabular}{l|c|c}
    \toprule
    \textbf{Variants} & \textbf{IoU} & \textbf{mIoU} \\
    \midrule
    \midrule
    HD$^2$-SSC (\textbf{Ours}) & \textbf{47.59} & \textbf{17.44} \\
    w/o $L_{\rm orth}$ & 46.93 (-0.66) & 16.64 (-0.8) \\
    w/o $L_{\rm decouple}$ & 46.85 (-0.74) & 16.78 (-0.66) \\
    w/o $L_{\rm critical}$ & 46.49 (-1.1) & 16.31 (-1.13) \\
    \bottomrule
    \end{tabular}
    \caption{Ablation study on the SemanticKITTI validation set dataset of different loss terms.}
    \label{loss}
\end{table}
\subsubsection{Implementation Details.}
We crop the input RGB images of cam2 to size $1220\times370$ for SemanticKITTI and images of cam1 to size $1408\times376$ for SSCBench-KITTI-360, extracting 2D feature maps with $1/16$ of input resolutions for subsequent process. The feature dimension $C$ is set to 128. The size $H\times W\times Z$ of the 3D feature volume is $128\times128\times16$, and the final predictions are upsampled to $256\times256\times32$. The channel dimensions for 2D pixel features and 3D voxel features are set to $C_{\rm 2D}=256$ and $C_{\rm 3D}=32$, respectively. The expanded dimension in the pseudo voxelization block is $D_{\rm exp}=4$, and the number of pixel and voxel query vectors is $N_{\rm query}=100$. The number of geometric and semantic critical voxels is set as $k=4096$. The above hyperparameters are practically selected by tuning on the validation set.
HD$^2$-SSC is trained for 24 epochs on 4 A6000 GPUs with a total batchsize of 4. The AdamW~\cite{loshchilov2017decoupled} optimizer is utilized with an initial learning rate of 2e-4 and a weight decay of 1e-2.
\begin{figure}[t]
\centering
\includegraphics[width=1.0\columnwidth]{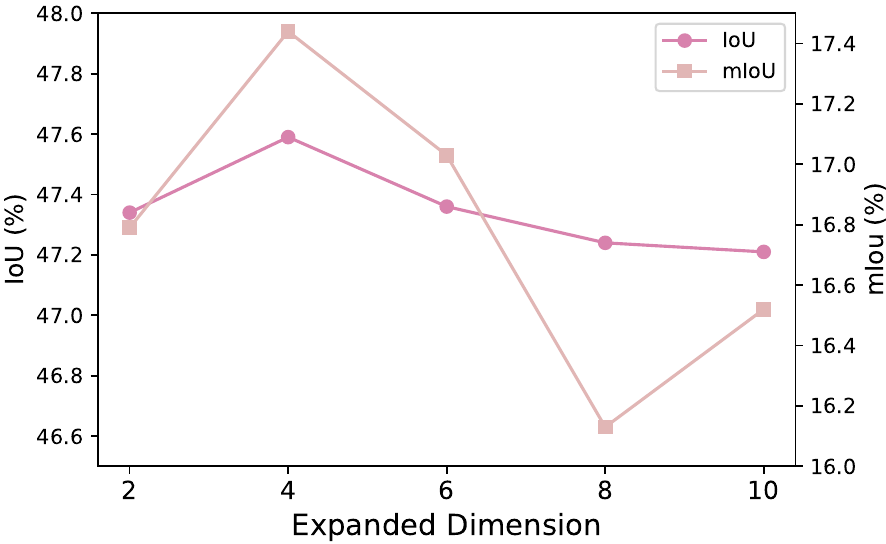} 
\caption{Effect of the expanded dimension on the SSC performance, evaluated on the SemanticKITTI validation set.}
\label{fig:exp}
\end{figure}
\begin{figure*}[t]
\centering
\includegraphics[width=\textwidth]{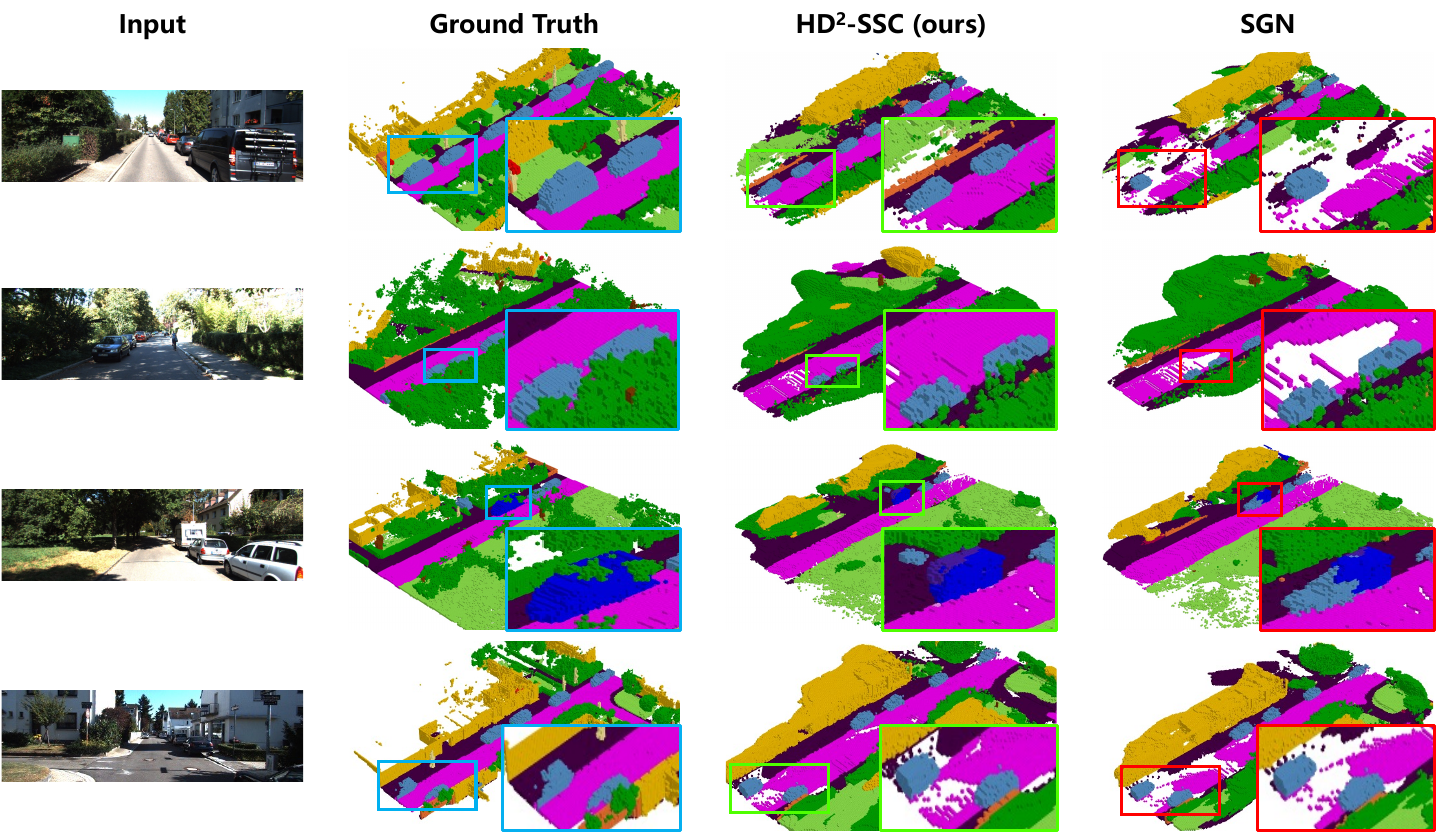} 
\caption{Visualization results of SSC prediction on the SemanticKITTI validation set. We highlight the occupancy ground truth with blue boxes, false SSC predictions of the best comparison method SGN with red boxes, and the improved SSC predictions from our HD$^2$-SSC approach with green boxes. Better viewed when zoomed in.}
\label{fig:vis}
\end{figure*}
\subsection{Main Results}
\subsubsection{Quantitative comparison.}
Table~\ref{kitti} and Table~\ref{kitti360} present the comparison results between our HD$^2$-SSC approach and other state-of-the-art camera-based SSC methods on the SemanticKITTI validation set and SSCBench-KITTI-360 test set respectively, where the best results are highlighted in bold and the second-best results are underlined. 
It can be observed that our HD$^2$-SSC approach achieves state-of-the-art performance in terms of both class-agnostic scene completion (IoU) and class-aware semantic scene completion (mIoU).
Specifically, regarding the best comparison method SGN~\cite{mei2024camera}, our approach achieves superior performance with \textbf{1.38}$\%$ IoU, \textbf{2.12}$\%$ mIoU enhancement on the SemanticKITTI validation set and \textbf{1.52}$\%$ IoU, \textbf{2.37}$\%$ mIoU enhancement on the SSCBench-KITTI-360 test set, respectively. Such performance improvements are attributed to the decoupled pixel semantics and aligned contextual details, which address the dimension and density gap. Although SGN employs a ``dense-sparse-dense" architecture for dynamically selecting discriminative voxel features, it overlooks the coarse and confusing pixel semantics in image features, and thus suffers from misaligned geometric and semantic contextual details. In contrast, our HD$^2$-SSC approach expands and decouples the coarse pixel semantics with orthogonal and decoupling loss, then aligns geometric and semantic critical voxels for consistent contextual details, enabling accurate SSC predictions with the completion of missing voxels and correction of erroneous ones.
\subsubsection{Ablation on architectural components.}
To further assess the effectiveness of HD$^2$-SSC, we conduct an ablation study on the SemanticKITTI validation set, evaluating the impact of key components. We incrementally integrate the HSD and HOR modules into the baseline method~\cite{li2023voxformer}, with results summarized in Table~\ref{ablation}.
Incorporating HSD yields improvements of 2.30\% in IoU and 2.23\% in mIoU over the baseline. Adding the HOR module enhances SSC performance over the baseline by 1.92\% in IoU and 2.77\% in mIoU. These results underscore the benefits of decoupling coarse pixel semantics and refining contextual details with the aligned geometric and semantic distribution. Notably, HSD and HOR modules complement each other, with their combined application achieving superior results.
\subsubsection{Ablation on different losses.} Table~\ref{loss} presents the ablation results on different losses, where disabling each loss causes performance degradation, validating their effectiveness:
\begin{itemize}
    \item $L_{\rm orth}$ improves the semantic diversity of expanded 2D features, decoupling coarse pixel semantics.
    \item $L_{\rm decouple}$ promotes semantic distinction in semantic clusters, enhancing the aggregated features.
    \item $L_{\rm critical}$ aligns critical voxel distributions, improving semantic and geometry consistency in the SSC prediction.
\end{itemize}

\subsubsection{Effect of expanded dimension.} Figure~\ref{fig:exp} demonstrates the effect of expanded dimension, where the best performance is achieved with $D_{\rm exp}=4$. When the expanded dimension keeps increasing, SSC performance begins to degrade. This phenomenon arises because, without explicit pixel semantic labels, too many expanded dimensions may bring additional confusion of ``imaginary" semantics that do not correspond to real-world objects, leading to inferior performance.
\subsubsection{Qualitative results.}
Figure~\ref{fig:vis} provides the visualization results from the SemanticKITTI validation set. To better illustrate the effectiveness of our HD$^2$-SSC approach, we also present the visualization results of the best comparison method SGN~\cite{mei2024camera} and the corresponding ground truth illustrations.
We outline the occupancy ground truth with blue boxes in the second row as a reference for comparison. The red boxes in the fourth row indicate the area of false voxel predictions where the SGN struggles, while the green boxes in the third row highlight the improved scene completion results from our HD$^2$-SSC approach with completion of missing voxels and correction of erroneous ones. This improvement is attributed to HD$^2$-SSC addressing the dimension and density gap with decoupled pixel semantics and refined voxel predictions.
\section{Conclusion}
In this paper, we identify the challenges of input-output dimension gap and annotation-reality density gap in camera-based SSC, and introduce the HD$^2$-SSC approach to address them. The HSD module expands and decouples coarse pixel semantics with orthogonal and decoupling loss, mitigating the dimension gap with the aggregation of voxelized semantics.
The HOR module aligns geometric and semantic critical voxels in a ``detect-and-refine" architecture, promoting consistent contextual details to bridge the density gap. Experiments and analyses on the SemanticKITTI and SSCBench-KITTI-360 datasets validate the effectiveness of our HD$^2$-SSC approach with superior SSC performance.
\\
\textbf{Limitation.} Despite our effectiveness, failure cases such as false prediction and incomplete boundary still occur in severe occlusion and distant areas with confusing features. Our future work aims to incorporate physical regularities to complement low-quality semantic features around those areas.

\section{Acknowledgments}
This work was supported by the grants from the National Natural Science Foundation of China (62525201, 62132001, 62432001) and Beijing Natural Science Foundation (L247006, L257005).

\bibliography{aaai2026}

@String{Computing = "Computing" }

@String{Computer = "{IEEE} Computer" }

@String{Springer = "Springer-Verlag" }

@ArtifactSoftware{R,
    title = {R: A Language and Environment for Statistical Computing},
    author = {{R Core Team}},
    organization = {R Foundation for Statistical Computing},
    address = {Vienna, Austria},
    year = {2019},
    url = {https://www.R-project.org/},
}

@String(ICCV= {Int. Conf. Comput. Vis.})

@String(ECCV= {Eur. Conf. Comput. Vis.})

@String(AAAI = {AAAI})

@String(ICCV  = {ICCV})

@String(ECCV  = {ECCV})

@article{liong2020amvnet,
  title={Amvnet: Assertion-based multi-view fusion network for lidar semantic segmentation},
  author={Liong, Venice Erin and Nguyen, Thi Ngoc Tho and Widjaja, Sergi and Sharma, Dhananjai and Chong, Zhuang Jie},
  journal={arXiv preprint arXiv:2012.04934},
  year={2020}
}

@inproceedings{cheng20212,
  title={2-s3net: Attentive feature fusion with adaptive feature selection for sparse semantic segmentation network},
  author={Cheng, Ran and Razani, Ryan and Taghavi, Ehsan and Li, Enxu and Liu, Bingbing},
  booktitle={Proceedings of the IEEE/CVF conference on computer vision and pattern recognition},
  pages={12547--12556},
  year={2021}
}

@inproceedings{ye2023lidarmultinet,
  title={Lidarmultinet: Towards a unified multi-task network for lidar perception},
  author={Ye, Dongqiangzi and Zhou, Zixiang and Chen, Weijia and Xie, Yufei and Wang, Yu and Wang, Panqu and Foroosh, Hassan},
  booktitle={Proceedings of the AAAI Conference on Artificial Intelligence},
  volume={37},
  number={3},
  pages={3231--3240},
  year={2023}
}

@inproceedings{cao2022monoscene,
  title={Monoscene: Monocular 3d semantic scene completion},
  author={Cao, Anh-Quan and de Charette, Raoul},
  booktitle={Proceedings of the IEEE/CVF Conference on Computer Vision and Pattern Recognition},
  pages={3991--4001},
  year={2022}
}

@inproceedings{li2023voxformer,
  title={Voxformer: Sparse voxel transformer for camera-based 3d semantic scene completion},
  author={Li, Yiming and Yu, Zhiding and Choy, Christopher and Xiao, Chaowei and Alvarez, Jose M and Fidler, Sanja and Feng, Chen and Anandkumar, Anima},
  booktitle={Proceedings of the IEEE/CVF Conference on Computer Vision and Pattern Recognition},
  pages={9087--9098},
  year={2023}
}

@article{li2023sscbench,
  title={SSCBench: A Large-Scale 3D Semantic Scene Completion Benchmark for Autonomous Driving},
  author={Li, Yiming and Li, Sihang and Liu, Xinhao and Gong, Moonjun and Li, Kenan and Chen, Nuo and Wang, Zijun and Li, Zhiheng and Jiang, Tao and Yu, Fisher and others},
  journal={arXiv preprint arXiv:2306.09001},
  year={2023}
}

@inproceedings{huang2023tri,
  title={Tri-perspective view for vision-based 3d semantic occupancy prediction},
  author={Huang, Yuanhui and Zheng, Wenzhao and Zhang, Yunpeng and Zhou, Jie and Lu, Jiwen},
  booktitle={Proceedings of the IEEE/CVF Conference on Computer Vision and Pattern Recognition},
  pages={9223--9232},
  year={2023}
}

@inproceedings{jiang2023polarformer,
  title={Polarformer: Multi-camera 3d object detection with polar transformer},
  author={Jiang, Yanqin and Zhang, Li and Miao, Zhenwei and Zhu, Xiatian and Gao, Jin and Hu, Weiming and Jiang, Yu-Gang},
  booktitle={Proceedings of the AAAI Conference on Artificial Intelligence},
  volume={37},
  number={1},
  pages={1042--1050},
  year={2023}
}

@article{huang2021bevdet,
  title={Bevdet: High-performance multi-camera 3d object detection in bird-eye-view},
  author={Huang, Junjie and Huang, Guan and Zhu, Zheng and Ye, Yun and Du, Dalong},
  journal={arXiv preprint arXiv:2112.11790},
  year={2021}
}

@inproceedings{li2023bevdepth,
  title={Bevdepth: Acquisition of reliable depth for multi-view 3d object detection},
  author={Li, Yinhao and Ge, Zheng and Yu, Guanyi and Yang, Jinrong and Wang, Zengran and Shi, Yukang and Sun, Jianjian and Li, Zeming},
  booktitle={Proceedings of the AAAI Conference on Artificial Intelligence},
  volume={37},
  number={2},
  pages={1477--1485},
  year={2023}
}

@article{tian2024occ3d,
  title={Occ3d: A large-scale 3d occupancy prediction benchmark for autonomous driving},
  author={Tian, Xiaoyu and Jiang, Tao and Yun, Longfei and Mao, Yucheng and Yang, Huitong and Wang, Yue and Wang, Yilun and Zhao, Hang},
  journal={Advances in Neural Information Processing Systems},
  volume={36},
  year={2024}
}

@article{mei2024camera,
  title={Camera-based 3d semantic scene completion with sparse guidance network},
  author={Mei, Jianbiao and Yang, Yu and Wang, Mengmeng and Zhu, Junyu and Ra, Jongwon and Ma, Yukai and Li, Laijian and Liu, Yong},
  journal={IEEE Transactions on Image Processing},
  year={2024},
  publisher={IEEE}
}

@inproceedings{yao2023ndc,
  title={Ndc-scene: Boost monocular 3d semantic scene completion in normalized device coordinates space},
  author={Yao, Jiawei and Li, Chuming and Sun, Keqiang and Cai, Yingjie and Li, Hao and Ouyang, Wanli and Li, Hongsheng},
  booktitle={2023 IEEE/CVF International Conference on Computer Vision (ICCV)},
  pages={9421--9431},
  year={2023},
  organization={IEEE Computer Society}
}

@inproceedings{wei2023surroundocc,
  title={Surroundocc: Multi-camera 3d occupancy prediction for autonomous driving},
  author={Wei, Yi and Zhao, Linqing and Zheng, Wenzhao and Zhu, Zheng and Zhou, Jie and Lu, Jiwen},
  booktitle={Proceedings of the IEEE/CVF International Conference on Computer Vision},
  pages={21729--21740},
  year={2023}
}

@inproceedings{zhang2023occformer,
  title={Occformer: Dual-path transformer for vision-based 3d semantic occupancy prediction},
  author={Zhang, Yunpeng and Zhu, Zheng and Du, Dalong},
  booktitle={Proceedings of the IEEE/CVF International Conference on Computer Vision},
  pages={9433--9443},
  year={2023}
}

@inproceedings{behley2019semantickitti,
  title={Semantickitti: A dataset for semantic scene understanding of lidar sequences},
  author={Behley, Jens and Garbade, Martin and Milioto, Andres and Quenzel, Jan and Behnke, Sven and Stachniss, Cyrill and Gall, Jurgen},
  booktitle={Proceedings of the IEEE/CVF international conference on computer vision},
  pages={9297--9307},
  year={2019}
}

@inproceedings{wang2024not,
  title={Not all voxels are equal: Hardness-aware semantic scene completion with self-distillation},
  author={Wang, Song and Yu, Jiawei and Li, Wentong and Liu, Wenyu and Liu, Xiaolu and Chen, Junbo and Zhu, Jianke},
  booktitle={Proceedings of the IEEE/CVF Conference on Computer Vision and Pattern Recognition},
  pages={14792--14801},
  year={2024}
}

@inproceedings{xue2024bi,
  title={Bi-SSC: Geometric-Semantic Bidirectional Fusion for Camera-based 3D Semantic Scene Completion},
  author={Xue, Yujie and Li, Ruihui and Wu, Fan and Tang, Zhuo and Li, Kenli and Duan, Mingxing},
  booktitle={Proceedings of the IEEE/CVF Conference on Computer Vision and Pattern Recognition},
  pages={20124--20134},
  year={2024}
}

@inproceedings{geiger2012we,
  title={Are we ready for autonomous driving? the kitti vision benchmark suite},
  author={Geiger, Andreas and Lenz, Philip and Urtasun, Raquel},
  booktitle={2012 IEEE conference on computer vision and pattern recognition},
  pages={3354--3361},
  year={2012},
  organization={IEEE}
}

@article{loshchilov2017decoupled,
  title={Decoupled weight decay regularization},
  author={Loshchilov, I},
  journal={arXiv preprint arXiv:1711.05101},
  year={2017}
}

@inproceedings{zou2021up,
  title={Up-to-down network: Fusing multi-scale context for 3d semantic scene completion},
  author={Zou, Hao and Yang, Xuemeng and Huang, Tianxin and Zhang, Chujuan and Liu, Yong and Li, Wanlong and Wen, Feng and Zhang, Hongbo},
  booktitle={2021 IEEE/RSJ International Conference on Intelligent Robots and Systems (IROS)},
  pages={16--23},
  year={2021},
  organization={IEEE}
}

@inproceedings{roldao2020lmscnet,
  title={Lmscnet: Lightweight multiscale 3d semantic completion},
  author={Roldao, Luis and de Charette, Raoul and Verroust-Blondet, Anne},
  booktitle={2020 International Conference on 3D Vision (3DV)},
  pages={111--119},
  year={2020},
  organization={IEEE}
}

@inproceedings{zhang2018efficient,
  title={Efficient semantic scene completion network with spatial group convolution},
  author={Zhang, Jiahui and Zhao, Hao and Yao, Anbang and Chen, Yurong and Zhang, Li and Liao, Hongen},
  booktitle={Proceedings of the European Conference on Computer Vision (ECCV)},
  pages={733--749},
  year={2018}
}

@inproceedings{cheng2021s3cnet,
  title={S3cnet: A sparse semantic scene completion network for lidar point clouds},
  author={Cheng, Ran and Agia, Christopher and Ren, Yuan and Li, Xinhai and Bingbing, Liu},
  booktitle={Conference on Robot Learning},
  pages={2148--2161},
  year={2021},
  organization={PMLR}
}

@article{rist2021semantic,
  title={Semantic scene completion using local deep implicit functions on lidar data},
  author={Rist, Christoph B and Emmerichs, David and Enzweiler, Markus and Gavrila, Dariu M},
  journal={IEEE transactions on pattern analysis and machine intelligence},
  volume={44},
  number={10},
  pages={7205--7218},
  year={2021},
  publisher={IEEE}
}

@inproceedings{xia2023scpnet,
  title={Scpnet: Semantic scene completion on point cloud},
  author={Xia, Zhaoyang and Liu, Youquan and Li, Xin and Zhu, Xinge and Ma, Yuexin and Li, Yikang and Hou, Yuenan and Qiao, Yu},
  booktitle={Proceedings of the IEEE/CVF conference on computer vision and pattern recognition},
  pages={17642--17651},
  year={2023}
}

@inproceedings{yangsemantic,
  title={Semantic segmentation-assisted scene completion for lidar point clouds},
  author={Yang, Xuemeng and Zou, Hao and Kong, Xin and Huang, Tianxin and Liu, Yong and Li, Wanlong and Wen, Feng and Zhang, Hongbo},
  booktitle={RSJ International Conference on Intelligent Robots and Systems (IROS)},
  pages={3555--3562},
    year={2021}
}

@inproceedings{mei2023ssc,
  title={SSC-RS: Elevate LiDAR semantic scene completion with representation separation and BEV fusion},
  author={Mei, Jianbiao and Yang, Yu and Wang, Mengmeng and Huang, Tianxin and Yang, Xuemeng and Liu, Yong},
  booktitle={2023 IEEE/RSJ International Conference on Intelligent Robots and Systems (IROS)},
  pages={1--8},
  year={2023},
  organization={IEEE}
}

@article{miao2023occdepth,
  title={Occdepth: A depth-aware method for 3d semantic scene completion},
  author={Miao, Ruihang and Liu, Weizhou and Chen, Mingrui and Gong, Zheng and Xu, Weixin and Hu, Chen and Zhou, Shuchang},
  journal={arXiv preprint arXiv:2302.13540},
  year={2023}
}

@inproceedings{tong2023scene,
  title={Scene as occupancy},
  author={Tong, Wenwen and Sima, Chonghao and Wang, Tai and Chen, Li and Wu, Silei and Deng, Hanming and Gu, Yi and Lu, Lewei and Luo, Ping and Lin, Dahua and others},
  booktitle={Proceedings of the IEEE/CVF International Conference on Computer Vision},
  pages={8406--8415},
  year={2023}
}

@inproceedings{li2022bevformer,
  title={Bevformer: Learning bird’s-eye-view representation from multi-camera images via spatiotemporal transformers},
  author={Li, Zhiqi and Wang, Wenhai and Li, Hongyang and Xie, Enze and Sima, Chonghao and Lu, Tong and Qiao, Yu and Dai, Jifeng},
  booktitle={European conference on computer vision},
  pages={1--18},
  year={2022},
  organization={Springer}
}

@inproceedings{philion2020lift,
  title={Lift, splat, shoot: Encoding images from arbitrary camera rigs by implicitly unprojecting to 3d},
  author={Philion, Jonah and Fidler, Sanja},
  booktitle={Computer Vision--ECCV 2020: 16th European Conference, Glasgow, UK, August 23--28, 2020, Proceedings, Part XIV 16},
  pages={194--210},
  year={2020},
  organization={Springer}
}

@inproceedings{jiang2024symphonize,
  title={Symphonize 3d semantic scene completion with contextual instance queries},
  author={Jiang, Haoyi and Cheng, Tianheng and Gao, Naiyu and Zhang, Haoyang and Lin, Tianwei and Liu, Wenyu and Wang, Xinggang},
  booktitle={Proceedings of the IEEE/CVF Conference on Computer Vision and Pattern Recognition},
  pages={20258--20267},
  year={2024}
}

@article{yao2023depthssc,
  title={Depthssc: Depth-spatial alignment and dynamic voxel resolution for monocular 3d semantic scene completion},
  author={Yao, Jiawei and Zhang, Jusheng},
  journal={arXiv preprint arXiv:2311.17084},
  year={2023}
}

@inproceedings{li2023lode,
  title={Lode: Locally conditioned eikonal implicit scene completion from sparse lidar},
  author={Li, Pengfei and Zhao, Ruowen and Shi, Yongliang and Zhao, Hao and Yuan, Jirui and Zhou, Guyue and Zhang, Ya-Qin},
  booktitle={2023 IEEE International Conference on Robotics and Automation (ICRA)},
  pages={8269--8276},
  year={2023},
  organization={IEEE}
}

@inproceedings{he2016deep,
  title={Deep residual learning for image recognition},
  author={He, Kaiming and Zhang, Xiangyu and Ren, Shaoqing and Sun, Jian},
  booktitle={Proceedings of the IEEE conference on computer vision and pattern recognition},
  pages={770--778},
  year={2016}
}

@inproceedings{lin2017feature,
  title={Feature pyramid networks for object detection},
  author={Lin, Tsung-Yi and Doll{\'a}r, Piotr and Girshick, Ross and He, Kaiming and Hariharan, Bharath and Belongie, Serge},
  booktitle={Proceedings of the IEEE conference on computer vision and pattern recognition},
  pages={2117--2125},
  year={2017}
}

@inproceedings{li2023bevstereo,
  title={Bevstereo: Enhancing depth estimation in multi-view 3d object detection with temporal stereo},
  author={Li, Yinhao and Bao, Han and Ge, Zheng and Yang, Jinrong and Sun, Jianjian and Li, Zeming},
  booktitle={Proceedings of the AAAI Conference on Artificial Intelligence},
  volume={37},
  number={2},
  pages={1486--1494},
  year={2023}
}

@inproceedings{wang2024h2gformer,
  title={H2gformer: Horizontal-to-global voxel transformer for 3d semantic scene completion},
  author={Wang, Yu and Tong, Chao},
  booktitle={Proceedings of the AAAI Conference on Artificial Intelligence},
  volume={38},
  number={6},
  pages={5722--5730},
  year={2024}
}

@inproceedings{ma2024cotr,
  title={Cotr: Compact occupancy transformer for vision-based 3d occupancy prediction},
  author={Ma, Qihang and Tan, Xin and Qu, Yanyun and Ma, Lizhuang and Zhang, Zhizhong and Xie, Yuan},
  booktitle={Proceedings of the IEEE/CVF Conference on Computer Vision and Pattern Recognition},
  pages={19936--19945},
  year={2024}
}

@article{zhang2023attention,
  title={Attention guided enhancement network for weakly supervised semantic segmentation},
  author={Zhang, Zhe and Wang, Bilin and Yu, Zhezhou and Zhao, Fengzhi},
  journal={Chinese Journal of Electronics},
  volume={32},
  number={4},
  pages={896--907},
  year={2023},
  publisher={CIE}
}

@inproceedings{lee2024semcity,
  title={Semcity: Semantic scene generation with triplane diffusion},
  author={Lee, Jumin and Lee, Sebin and Jo, Changho and Im, Woobin and Seon, Juhyeong and Yoon, Sung-Eui},
  booktitle={Proceedings of the IEEE/CVF conference on computer vision and pattern recognition},
  pages={28337--28347},
  year={2024}
}

@article{roldao20223d,
  title={3D semantic scene completion: A survey},
  author={Roldao, Luis and De Charette, Raoul and Verroust-Blondet, Anne},
  journal={International Journal of Computer Vision},
  volume={130},
  number={8},
  pages={1978--2005},
  year={2022},
  publisher={Springer}
}

@article{yu2024context,
  title={Context and geometry aware voxel transformer for semantic scene completion},
  author={Yu, Zhu and Zhang, Runmin and Ying, Jiacheng and Yu, Junchen and Hu, Xiaohai and Luo, Lun and Cao, Si-Yuan and Shen, Hui-Liang},
  journal={arXiv preprint arXiv:2405.13675},
  year={2024}
}

@article{huang2024probabilistic,
  title={Probabilistic Gaussian Superposition for Efficient 3D Occupancy Prediction},
  author={Huang, Yuanhui and Thammatadatrakoon, Amonnut and Zheng, Wenzhao and Zhang, Yunpeng and Du, Dalong and Lu, Jiwen},
  journal={arXiv preprint arXiv:2412.04384},
  year={2024}
}

@inproceedings{tan2019efficientnet,
  title={Efficientnet: Rethinking model scaling for convolutional neural networks},
  author={Tan, Mingxing and Le, Quoc},
  booktitle={International conference on machine learning},
  pages={6105--6114},
  year={2019},
  organization={PMLR}
}

@article{xiao2024instance,
  title={Instance-aware monocular 3D semantic scene completion},
  author={Xiao, Haihong and Xu, Hongbin and Kang, Wenxiong and Li, Yuqiong},
  journal={IEEE Transactions on Intelligent Transportation Systems},
  volume={25},
  number={7},
  pages={6543--6554},
  year={2024},
  publisher={IEEE}
}

@article{huang2022bevdet4d,
  title={Bevdet4d: Exploit temporal cues in multi-camera 3d object detection},
  author={Huang, Junjie and Huang, Guan},
  journal={arXiv preprint arXiv:2203.17054},
  year={2022}
}

@article{wang2025l2cocc,
  title={L2COcc: Lightweight Camera-Centric Semantic Scene Completion via Distillation of LiDAR Model},
  author={Wang, Ruoyu and Ma, Yukai and Yao, Yi and Tao, Sheng and Li, Haoang and Zhu, Zongzhi and Liu, Yong and Zuo, Xingxing},
  journal={arXiv preprint arXiv:2503.12369},
  year={2025}
}

@inproceedings{yang2025sphere,
  title={SPHERE: Semantic-PHysical Engaged REpresentation for 3D Semantic Scene Completion},
  author={Yang, Zhiwen and Peng, Yuxin},
  booktitle={Proceedings of the 33rd ACM International Conference on Multimedia},
  pages={7681--7690},
  year={2025}
}

@article{yang2025gala,
  title={GaLa-2.5 D: Global-Local Alignment with 2.5 D Semantic Guidance for Camera-based 3D Semantic Scene Completion in Autonomous Driving},
  author={Yang, Zhiwen and Peng, Yuxin},
  journal={Chinese Journal of Electronics},
  year={2025}
}

\clearpage
\appendix

\section{Appendix}

\subsection{Efficiency Analysis}
Table~\ref{tab:efficiency} below presents the efficiency evaluation results of our HD$^2$-SSC approach against the best comparison method SGN~\cite{mei2024camera}. It can be observed that HD$^2$-SSC achieves superior performance with slightly increased model parameters and GFLOPs, even less GPU memory and inference time, since our enhanced feature enables accurate SSC prediction with (128,128,16) feature grids, avoiding SGN's costly up-sampling to (256,256,32) grids.

\begin{table}[h]
    \centering
    \caption{Efficiency evaluation against the best comparison method SGN on 1 NVIDIA A6000 GPU.}
    \resizebox{1.0\columnwidth}{!}{
    \begin{tabular}{c|cccc}
    \toprule
    \textbf{Methods} & \textbf{Params~(M)} & \textbf{Memory~(G)} & \textbf{GFLOPs} & \textbf{Inference~(s)} \\
    \midrule
    \midrule
    SGN & 28.16 & 15.83 & 725.05 & 0.61 \\
    \rowcolor{gray!10} 
    \textbf{HD$^2$-SSC}& 28.96 & 14.42 & 842.76 & 0.56 \\
    \bottomrule
    \end{tabular}
    }
    \label{tab:efficiency}
\end{table}

\subsection{Dataset Generalization}
Table~\ref{tab:nuScenes} below presents the evaluation results on the Occ3D-nuScenes~\cite{tian2024occ3d} dataset, where our HD$^2$-SSC outperforms popular OccFormer~\cite{zhang2023occformer} and BEVDet4D~\cite{huang2022bevdet4d} baselines, demonstrating our generalizability across different datasets.

\begin{table}[h]
    \centering
    \caption{Evaluation on the Occ3D-nuScenes dataset.}
    \begin{tabular}{c|cc}
    \toprule
    \textbf{Methods} & \textbf{IoU} & \textbf{mIoU} \\
    \midrule
    \midrule
    OccFormer & 70.1 & 37.4 \\
    BEVDet4D & 73.8 & 39.3 \\
    \rowcolor{gray!10} 
    \textbf{HD$^2$-SSC~(Ours)}& \textbf{75.4} & \textbf{44.2} \\
    \bottomrule
    \end{tabular}
    \label{tab:nuScenes}
\end{table}

\subsection{Evaluation with EfficientNet-B7}
Table~\ref{tab:efficientnet} presents the evaluation results with EfficientNet-B7~\cite{tan2019efficientnet} backbone on the SSCBench-KITTI-360~\cite{li2023sscbench} test set, where our HD$^2$-SSC achieves superior performance compared to CGForemr~\cite{yu2024context} and L2COcc~\cite{wang2025l2cocc}, further demonstrating our effectiveness in bridging the dimension and density gaps in the SSC task.

\begin{table}[h]
    \centering
    \caption{Evaluation with EfficientNet-B7 backbone on the SSCBench-KITTI-360 test set.}
    \begin{tabular}{c|cc}
    \toprule
    \textbf{Methods} & \textbf{IoU} & \textbf{mIoU} \\
    \midrule
    \midrule
    CGFormer & 48.07 & 20.05 \\
    L2COcc & 48.07 & 20.11 \\
    \rowcolor{gray!10} 
    \textbf{HD$^2$-SSC~(Ours)}& \textbf{49.03} & \textbf{21.13} \\
    \bottomrule
    \end{tabular}
    \label{tab:efficientnet}
\end{table}

\subsection{Failure Case Analysis}
\begin{figure}[t]
\centering
\includegraphics[width=1.0\columnwidth]{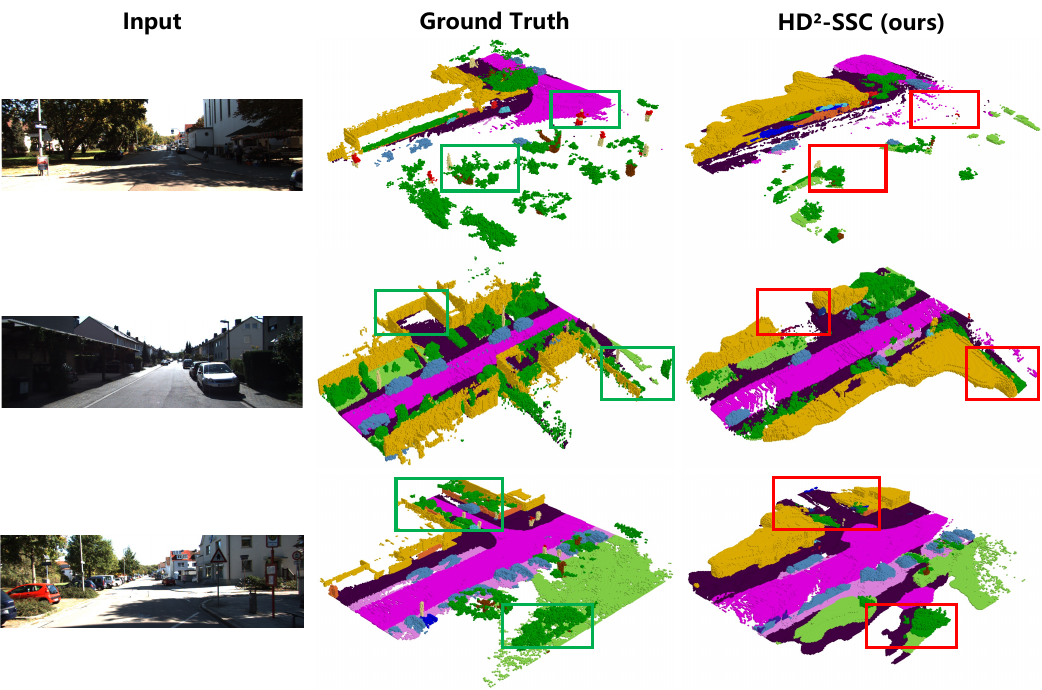}
\caption{Failure case analysis on the SemanticKITTI validation set. We highlight the occupancy ground truth with green boxes, false SSC predictions of our HD$^2$-SSC approach with red boxes. Better viewed when zoomed in.}
\label{fig:fail}
\end{figure}

Figure~\ref{fig:fail} presents the failure case analysis on the SemanticKITTI validation set, where failure cases such as false occupancy prediction and incomplete object boundary occur in severe occlusion and distant areas due to the low-quality features with confusing semantics. Our future work aims to incorporate physical regularities to complement low-quality semantic features around those areas.

\end{document}